\documentclass[journal]{IEEEtran}
\usepackage{cite}
\usepackage{bbm}
\usepackage{url}
\usepackage{xcolor}
\ifCLASSINFOpdf
\else
\fi
\usepackage{graphicx}
\usepackage{amsmath, bm}
\usepackage{array}

\usepackage{algpseudocode}
\usepackage{algorithm}

\begin{document}
\title{Fitting Multiple Machine Learning Models with Performance Based Clustering} 
\author{Mehmet E. Lorasdagi, Ahmet B. Koc, Ali T. Koc, and Suleyman S. Kozat, \IEEEmembership{Senior Member, IEEE}
\thanks{© 2025 IEEE. Personal use of this material is permitted. Permission from IEEE must be 
obtained for all other uses, in any current or future media, including 
reprinting/republishing this material for advertising or promotional purposes, creating new 
collective works, for resale or redistribution to servers or lists, or reuse of any copyrighted 
component of this work in other works.}
\thanks{This study is supported by Turk Telekom within the framework of 5G and Beyond Joint Graduate Support Programme coordinated by Information and Communication Technologies Authority.}}

\markboth{IEEE SIGNAL PROCESSING LETTERS}%
{Shell \MakeLowercase{\textit{et al.}}: Bare Demo of IEEEtran.cls for IEEE Journals}
\maketitle
\begin{abstract}
Traditional machine learning approaches assume that data comes from a single generating mechanism, which may not hold for most real life data. In these cases, the single mechanism assumption can result in suboptimal performance. We introduce a clustering framework that eliminates this assumption by grouping the data according to the relations between the features and the target values, and we obtain multiple separate models to learn different parts of the data. We further extend our framework to applications having streaming data where we produce outcomes using an ensemble of models. For this, the ensemble weights are updated based on the incoming data batches. We demonstrate the performance of our approach over the widely-studied real life datasets, showing significant improvements over the traditional single-model approaches. 
\end{abstract}

\IEEEpeerreviewmaketitle

\vspace{-0.6cm}
\section{Introduction}
\IEEEPARstart{I}{n} the signal processing and machine learning literature, it is generally assumed that data comes from a single generating mechanism, and machine learning (ML) models are developed under this assumption. However, this assumption might not hold in general, especially in real life applications due to nonstationarity, or possible different sources of data. For example, for the wind turbine energy production data, the energy generated at different wind farms can vary for the same wind speed and angle due to varying turbine dynamics \cite{wind_kaggle}. Another example is the spatiotemporal crime prediction data, where the relations between the features and the crime rate may vary based on the location and time \cite{crime_kaggle}. 

The single generating mechanism assumption causes ML models to learn an average of different parts of the data. This can cause suboptimal performance as the differences in separate relations are neglected. Using separate ML models to learn the different parts of the data can improve the data modelling performance. Vanilla clustering algorithms cluster the data based on only using the feature vectors, i.e., data points with similar feature structures are grouped together \cite{clustering1, clustering2}. Recent work \cite{VWKM} assigns feature-specific weights based on their significance to individual clusters. Furthermore, encoder-decoder based deep learning models \cite{SDCN}, \cite{learningrepresentation}, \cite{SimCLR} and \cite{DeepCluster} learn by transforming data to strong representations, then applying the  clustering algorithms on them.  Similarly,  RNN-based models \cite{DTC} and \cite{deepcl} learn dynamics and reduce dimensionality for sequential data. However, these approaches do not consider the relations between the features and the target values while constructing these clusters. These relations should be carefully preserved while clustering.

To this end, we introduce a clustering algorithm that works not on the features but on the relations between the feature vectors and the target data. Since the number of different generating mechanisms and their structure are unknown, we use an approach inspired by Expectation-Maximization (EM), where the cost function measures the distance of a data point to a function, to solve the optimization problem \cite{em}. By clustering based on the relations, our algorithm eliminates the single mechanism assumption and preserves the different relations between the feature vectors and the target data. 

We also investigate online or sequential applications where we observe the data in a streaming manner, i.e., the target values are revealed after the feature vectors are received, which is common in the deployment of ML algorithms in real life applications. To address this, we introduce an ensemble approach where we use a weighted average of the ML models obtained during clustering to provide outputs in a sequential manner. We use a gradient descent based learning approach to update the ensemble weights, where the ensemble weights of the functions are updated based on the incoming data batches. This approach is suitable for online modelling where temporally close data is likely to come from similar generating mechanisms \cite{ts1}. 

Our contributions are: 1) We introduce a performance based clustering framework that clusters the data based on the relations between the features and the target values. 2) We introduce an ensemble based online or sequential approach optimized for real life applications of the ML algorithms. 3) Through an extensive set of experiments over the well-known competition datasets, we illustrate significant performance gains compared to the vanilla models in sequential inference settings. 4) For the reproducibility of our results and to facilitate further research, we openly share our source code\footnote{\url{https://github.com/mefe06/function-clustering}}.
\vspace{-0.9cm}
\section{Problem Definition} \label{sec:problem_def}
Consider a dataset\footnote{In this letter, column vectors are represented using the bold lowercase letters, matrices are denoted using the bold uppercase letters. For a vector $ \bm{x} $ and a matrix $\bm{X}$, their respective transposes are represented as $ \bm{x}^T $ and $ \bm{X}^T $. ${x}_{k}$ represents the $k^{th}$ element of $\bm{x}$. The symbol $ \sum(\cdot) $ denotes the summation of elements in a vector or a matrix. $\mathbbm{1}_{\{\cdot\}}$ denotes the indicator function, which is 1 if the condition is satisfied and 0 otherwise.} $\mathcal{D}$. We assume $\mathcal{D}= \{\mathcal{D}_k\}^K_{k=1}$, where $\mathcal{D}_k$ is a set of data that contains $\{\bm{x}_i, y_i\}$ pairs with,
\begin{equation} \label{eq:generating_relation}
    y_i = f_k(\bm{x}_i)+ \epsilon \; \text{,}
\end{equation}
where $\bm{x}_i$ is the feature vector and $y_i$ is the corresponding target variable, $f_k(\cdot)$ is the function that generates the data for $\mathcal{D}_k$, $k \in \{1,.., K\}$, and $\epsilon$ is independent and identically distributed random noise. We denote the group membership of each pair $\{\bm{x}_i, y_i\}$ by $\gamma_i$, i.e., $\gamma_i = k$ in (\ref{eq:generating_relation}). We do not know the group memberships $\gamma_i$ of the pairs $\{\bm{x}_i, y_i\}$ in $\mathcal{D}$, i.e., which function generates a data point, and the generating functions $f_k(\cdot)$'s are unknown. For theoretical analysis, we assume $K$ is known and we assume no such prior knowledge in real life applications.

Our goal is to minimize the total modelling cost,
\begin{equation} \label{eq:obj}
    L(\hat{\bm{y}}, \bm{y}) = \frac{1}{N}\sum_{k=1}^{\hat{K}}\sum_{i=1}^{N}  l(\hat{f}_k(\bm{x_i}), y_i)\mathbbm{1}_{\{\bm{x}_i, y_i\} \in \hat{\mathcal{D}}_k} \; \text{,}
\end{equation}
where $N$ is the number of data points, $\hat{\bm{y}} = [\hat{f}_{\hat{\gamma}_1}(\bm{x_1}), ..., \hat{f}_{\hat{\gamma}_N}(\bm{x_N})]^T$ is our prediction of the true target vector $\bm{y} = [y_1, ..., y_N]^T$, $\hat{\gamma}_i$ is our prediction of the true group membership $\gamma_i$, $\hat{f}_k(\cdot)$ is any learned model of the true generating function $f_k(\cdot)$, $\hat{\mathcal{D}}_k$ is our prediction of the true data group ${\mathcal{D}}_k =  \{ \{\bm{x}_i, y_i\} \mid {\gamma}_i = k  \}$, $\hat{K}$ is our prediction of $K$, and $l(\cdot)$ can be any differentiable loss function. Note that neither $f_k(\cdot)$'s nor $\mathcal{D}_k$'s are known to us. 
 
We first introduce a performance based clustering framework to learn the true functions and their clusters. After learning the clusters, we use the corresponding models in an online approach to extend our framework to sequential settings.
\vspace{-0.5cm}
\section{Performance Based Clustering Approach} \label{sec:sol}
\subsection{Performance Based Clustering} \label{ssec:fc}
We use an EM inspired algorithm to solve the joint cluster identification and function learning problem in (\ref{eq:obj}) \cite{em}. We denote an ML model as $f(\cdot, {\bm{\theta}})$, where $\bm{\theta}$ are the parameters of the model. Note that any well-known ML algorithm(s) with sufficient learning power can be used accordingly.

We first initialize the clusters $\hat{\mathcal{D}}_k$, where $k \in \{1, ..., \hat{K}\}$. Note that $\hat{K}$ is selected using cross-validation in our simulations since it is also unknown. We first form an initial cluster by randomly sampling data points from the entire dataset $\mathcal{D}$, and fit a model over the points in the cluster. To initialize the remaining $\hat{K}-1$ clusters, we employ a specific distance-based selection method to ensure that new clusters capture data points that are poorly represented by the existing clusters. Specifically, for each unassigned data point, we compute the prediction errors under all the models of the already initialized clusters and define the distance of the data point as the best performance across these models. We then set the selection probabilities of the unassigned data points proportional to these distances. Thus, data points with higher prediction errors, i.e., those poorly explained by the existing models, have higher probabilities of being selected for the initialization of the next cluster. This approach ensures that each newly initialized cluster captures distinct regions of the relation space, i.e., the $\mathcal{D}_k$'s and the $f_k(\cdot)$'s. After selecting the points for the newly initialized cluster, we fit a new model to minimize the error over these data points.

After initializing the clusters, we iteratively update the clusters and their corresponding models to optimize the modelling performance. At the $m^{\text{th}}$ iteration, we first assign data points to the closest cluster as the expectation step, which corresponds to lines 6-9 in Algorithm \ref{alg:fc}. We define $\hat{{\bm{\theta}}}^{m}_k$ as our estimate of the parameters of the ML model for the $k^{\text{th}}$ cluster at the $m^{\text{th}}$ iteration.
We measure the distance $c_{ik}^m$ of the data point $\{\bm{x}_i, y_i\}$ to the cluster $\hat{\mathcal{D}}_k^{m-1}$ with the corresponding function $f(\cdot, {\hat{{\bm{\theta}}}}^{m-1}_k)$ as $c_{ik}^m = l(f(\bm{x}_i, {\hat{{\bm{\theta}}}}^{m-1}_k), y_i) \text{.} $
Then, we predict the new cluster assignment $\hat{\gamma}_i^m$ for the corresponding data point at the $m^{\text{th}}$ iteration
\begin{equation}\label{eq: max}
    \hat{\gamma}_i^m = \underset{k}{argmin} \; c_{ik}^m 
\end{equation}
and obtain the updated clusters $\hat{\mathcal{D}}_1^m, ..., \hat{\mathcal{D}}_{\hat{K}}^m$. Here, $ 
    \hat{\mathcal{D}}_k^m = \{ \{\bm{x}_i, y_i\} \mid \hat{\gamma}_i^m = k  \} \text{.} $ Note that we can also use soft assignments, where the probability of assigning $\{\bm{x}_i, y_i\}$ to the $k^{\text{th}}$ cluster is proportional to $e^{-c_{ik}^m}$. 
Then, for each cluster $\hat{\mathcal{D}}_k^m$ independently, we train an ML model ${f}(\cdot, {\bm{\theta}_k})$ to minimize the error over the current data points in the cluster,
\begin{equation} \label{eq:exp}
    {\hat{{\bm{\theta}}}}^m_k = \underset{{\bm{\theta}}_k}{argmin} \sum_{i=1}^N l(f(\bm{x}_i, {\bm{\theta}}_k), y_i)\mathbbm{1}_{\{\bm{x}_i, y_i\} \in \hat{\mathcal{D}}_k^m} \; \text{.}
\end{equation}
This is the maximization step in our framework, which corresponds to lines 10-12 in Algorithm \ref{alg:fc}. 

We apply these expectation and maximization steps until the reassignment process converges, i.e., using a convergence criterion as
\begin{equation} \label{eq:convergence}
    \frac{1}{N}\sum_{i=1}^{N}\mathbbm{1}_{\{\hat{\gamma}_i^m\neq \hat{\gamma}_i^{m-1}\}} < \zeta \; \text{,}
\end{equation}
where $\zeta$ is a predetermined constant, which is optimized using the validation set as shown in our simulations. Note that different convergence criteria can be used \cite{early_stop}.
\begin{algorithm}
\caption{Performance Based Clustering}
\label{alg:fc}
\begin{algorithmic}[1]
\State Initialize clusters $\hat{\mathcal{D}}_{1}^{0}, ..., \hat{\mathcal{D}}_{\hat{K}}^0$ 
\State Initialize models $f(\cdot, {\hat{{\bm{\theta}}}}^{0}_1), ..., f(\cdot, {\hat{{\bm{\theta}}}}^{0}_{\hat{K}})$
\State $m = 1 $
\While{stop criterion (\ref{eq:convergence}) not met}
    \For{each $\{\bm{x}_i, y_i\}$}
        \State $\hat{\gamma}_i^m = \underset{k}{argmin} \; l({f}(\bm{x}_i, {\hat{\bm{\theta}}}_k^{m-1}), y_i)$
        \State Assign $\{\bm{x}_i, y_i\}$ to cluster $\hat{\mathcal{D}}^m_{\hat{\gamma}_i^m}$
    \EndFor
    \For{$k = 1$ to $\hat{K}$}
        \State ${\hat{{\bm{\theta}}}}_k^m = \underset{{\bm{\theta}}_k}{argmin} \sum_{i=1}^N l(f(\bm{x}_i, {\bm{\theta}}_k), y_i)\mathbbm{1}_{\{\bm{x}_i, y_i\} \in \hat{\mathcal{D}}_k^m}$
    \EndFor
    \State $m=m+1$
\EndWhile
\State Set $\hat{\mathcal{D}}_k = \hat{\mathcal{D}}_k^{m-1}$ and $ \hat{\bm{\theta}}_k = \hat{\bm{\theta}}_k^{m-1} $ for $k =\{1, ..., \hat{K}\}$
\State \Return clusters $\hat{\mathcal{D}}_1, ..., \hat{\mathcal{D}}_{\hat{K}}$, models $f(\cdot, {\hat{{\bm{\theta}}}}_1), ..., f(\cdot,{\hat{{\bm{\theta}}}}_{\hat{K}})$
\end{algorithmic}
\end{algorithm}

Further, an iteration of this algorithm results in non-increasing loss, i.e., $L^m \leq L^{m-1}$, where $L^m$ denotes the total modelling loss at the $m^{\text{th}}$ iteration. Clearly,
\begin{equation}
    L^{m-1} = \frac{1}{N}\sum_{k=1}^{\hat{K}}\sum_{i=1}^N l(f(\bm{x}_i, {\hat{{\bm{\theta}}}}^{m-1}_k), y_i)\mathbbm{1}_{\{\bm{x}_i, y_i\} \in \hat{\mathcal{D}}_k^{m-1}} \; \text{.}
\end{equation}
We first obtain the new cluster assignments from the expectation step. For an arbitrary data point $\{\bm{x}_i, y_i\}$,
\begin{equation}
        l(f(\bm{x}_i, {\hat{{\bm{\theta}}}}^{m-1}_{\hat{\gamma}_i^m}), y_i) \leq l(f(\bm{x}_i, {\hat{{\bm{\theta}}}}^{m-1}_j), y_i) 
\end{equation}
from the definition of $\hat{\gamma}_i^m$ in (\ref{eq: max}). Then, 
\begin{equation}
    \frac{1}{N} \sum_{i=1}^N l(f(\bm{x}_i, {\hat{{\bm{\theta}}}}^{m-1}_{\hat{\gamma}_i^m}), y_i) \leq \frac{1}{N} \sum_{i=1}^N l(f(\bm{x}_i, {\hat{{\bm{\theta}}}}^{m-1}_{j_i}), y_i) \; \text{,}
\end{equation}
where $j_i \in \{1, ..., \hat{K}\}$, i.e., any arbitrary cluster. Thus, if we choose these arbitrary $j_i$'s to be $\hat{\gamma}_i^{m-1}$ for each data point, the loss is non-increasing in the expectation step,
\begin{equation}
    \frac{1}{N} \sum_{i=1}^N l(f(\bm{x}_i, {\hat{{\bm{\theta}}}}^{m-1}_{\hat{\gamma}_i^m}), y_i) \leq \frac{1}{N} \sum_{i=1}^N l(f(\bm{x}_i, {\hat{{\bm{\theta}}}}^{m-1}_{\hat{\gamma}_i^{m-1}}), y_i) \; \text{.}
\end{equation}
In the maximization step,
\begin{equation}
    \sum_{i=1}^N l(f(\bm{x}_i, {\hat{{\bm{\theta}}}}^m_{\hat{\gamma}_i^m}), y_i) \leq \sum_{i=1}^N l(f(\bm{x}_i, {\hat{{\bm{\theta}}}}^{m-1}_{\hat{\gamma}_i^m}), y_i) \; \text{,}
\end{equation}
where $f(\cdot, {\hat{{\bm{\theta}}}}^m_{\hat{\gamma}_i^m})$ is obtained according to (\ref{eq:exp}). This inequality follows from the definition of $f(\cdot, {\hat{{\bm{\theta}}}}^m_{\hat{\gamma}_i^m})$.

Thus, the maximization step does not increase the total loss from the expectation step.
Finally, 
\begin{equation}
    L^m = \frac{1}{N} \sum_{i=1}^N l({f}(\bm{x}_i, {\hat{\bm{\theta}}}^m_{\hat{\gamma}_i^m}), y_i) \leq L^{m-1} \; \text{.}
\end{equation}

Consider the sequence of the modelling losses, $L^m$'s, which are shown to be monotonically decreasing. Further, assume we chose perfect clusters, i.e., each data point is a member of the cluster whose function approximates the true generating mechanism perfectly. Assume we have unbiased estimators and ML models with the universal approximation property \cite{univ_approx}. Then, when $l(\cdot)$ is the squared error, the theoretical lower bound for the mean squared error (MSE) is the variance of the noise for each cluster from the Cramer-Rao bound \cite{cr_bound}. Then, the total minimum cost is a weighted average of the noise variances. Thus, there is a lower bound on the prediction error due to the noise. For other loss functions, we can find a lower bound on the variance of the estimator with the Cramer-Rao or the Chapman-Robbins bounds \cite{chap-rob}. Then, we can use the relation between the loss and the variance of the estimator to derive lower bounds, e.g. $E[|\hat{\theta} - \theta|] \geq c \cdot (\text{Var}(\hat{\theta}))^{\frac{1}{2}}
$ for the absolute error under certain assumptions on $\epsilon$. Hence, considering we have a monotonic bounded sequence, the sequence converges to its minima from the Monotone Convergence Theorem \cite{mct}. 

While the performance based clustering approach models offline data by identifying the different parts of the data, many real-world applications require sequential processing where the true outcome is revealed, if any, after we produce the outcome. To address this, we extend our method to online settings in the following section.
\vspace{-0.5 cm}
\subsection{Online or Sequential Approach} \label{ssec:streaming}
When the data arrives in a streaming manner, e.g., real life prediction problems with time series data, the observed features do not directly provide the class membership \cite{assignment}. Hence, in inference, we need additional assumptions to predict the cluster memberships, as we do not observe the target values while producing our outcomes. Here, we assume that the data produced by similar generating mechanisms arrive in temporally close batches. For this setting, we use a weighted ensemble of the outcomes of the models obtained in the offline setting in Section \ref{ssec:fc}.

We first uniformly initialize the ensemble weights $\bm{w}^0 = [w^0_1, ..., w^0_{\hat{K}}]^T$ at the start. Then, we produce outcomes with the current weights and the already learned functions for the incoming data batch $\mathcal{B}^t$. Next, we update the ensemble weights with gradient descent after observing the corresponding target values for the current batch.

The weight updates are derived as follows. At time $t$, the batch $\mathcal{B}^t = \{\bm{x}^{t}_i, y^{t}_i\}_{i=1}^{N_t}$ arrives without the target values $y^{t}_i$'s initially, i.e., only the feature vectors $\bm{x}^{t}_i$'s arrive. The class memberships are unknown, however, we assume the data points in the same batch come from the same class. Note that this assumption holds in many real life scenarios such as in renewable energy prediction, since the data generating mechanisms change slowly in time.

Then, for each $\bm{x}_i^{t}$ in the batch, we predict the corresponding target value as $
    \hat{y}^{t}_i = ({\bm{w}^{t-1}})^T\hat{\bm{g}}(\bm{x}_i^{t})\text{,} $
where $\hat{\bm{g}}(\bm{x}_i^{t}) = [{f}(\bm{x}_i^{t}, {\hat{\bm{\theta}}}_1), ..., {f}(\bm{x}_i^{t}, {\hat{\bm{\theta}}}_{\hat{K}})]^T$, ${\bm{w}}^{t-1} = [{w}^{t-1}_1, ..., {w}^{t-1}_{\hat{K}}]^T$. We define $\hat{\bm{y}}^{t} = [\hat{y}^{t}_1, ..., \hat{y}^{t}_{N^t}]^T$, where $N^t$ is the number of samples in $\mathcal{B}^t$. 

After our predictions, we observe the target values $\bm{y}^{t}=[y^{t}_1, ..., y^{t}_{N^t}]^T$. Then, the prediction error for the batch is
\begin{equation}
    L(\hat{\bm{y}}^{t}, \bm{y}^{t}) = \frac{1}{N^t}\sum_{i=1}^{N^t} l(\hat{y}^{t}_i, y^{t}_i) \; \text{.}
\end{equation}
We take the gradient of the prediction error with respect to the weight vector. Hence, we have:
\begin{equation}
        \frac{\partial L(\hat{\bm{y}}^{t}, \bm{y}^{t})}{\partial \bm{w}^{t-1}} =\frac{\partial L(\hat{\bm{y}}^{t}, \bm{y}^{t})}{\partial \hat{\bm{y}}^{t}}\frac{\partial \hat{\bm{y}}^{t}}{\partial \bm{w}^{t-1}} \; \text{.}
\end{equation}
Then, considering $l(\cdot)$ to be the squared error,   
\begin{equation} \label{eq: grad}
    \frac{\partial L(\hat{\bm{y}}^{t}, \bm{y}^{t})}{\partial \bm{w}^{t-1}} = \frac{2}{N^t}[\hat{\bm{g}}(\bm{x}_1^{t}), ...,\hat{\bm{g}}(\bm{x}_{N^t}^{t})] (\hat{\bm{y}}^{t} - \bm{y}^{t}) \; \text{.}
\end{equation}
We update the weights with gradient descent,
\begin{equation}
    \bm{w}^t = \bm{w}^{t-1} - \alpha \frac{2}{N^t}[\hat{\bm{g}}(\bm{x}_1^{t}), ..., \hat{\bm{g}}(\bm{x}_{N^t}^{t})] (\hat{\bm{y}}^{t} - \bm{y}^{t}) \; \text{,}
\end{equation}
where $\alpha$ is the learning rate hyperparameter, which can be optimized with validation.

We explain our experiments with the introduced approach and our discussions in the next section.
\vspace{-0.45cm}
\section{Experiments} \label{sec:experiments}

Here, we demonstrate the effectiveness of the introduced algorithm over both the synthetic data, under controlled settings, and the well-known real life competition datasets. We first start with the synthetic data and then continue with the real life data. In all experiments, we use Multi-Layer Perceptron (MLP) as our ML model, due to its impressive performance in real life competitions \cite{dl}.
\vspace{-0.45cm}
\subsection{Experiments Under Controlled Settings}
We begin with the synthetic data generation procedure, followed by the experiments under this controlled setting.

We generate the data using three different linear relations, i.e., $K=3$. For each relation $r \in \{1, 2, 3\}$, we generate $
    y_i^{(r)} = \sum_{j=1}^3 \beta_j^{(r)} x_{ij}^{(r)} + \epsilon_i^{(r)}$ 
, where $\beta_j^{(r)}$ is the randomly selected coefficient, $x_{ij}^{(r)} \sim \mathcal{N}(0, 1)$ is the feature value, and $\epsilon_i^{(r)} \sim \mathcal{N}(0, 0.1)$ is the Gaussian noise. We run 25 simulations and generate $N =5000$ different data points for each simulation, with approximately equal numbers from each relation.
\setlength{\textfloatsep}{5pt}
\begin{figure}
    \centering
    \includegraphics[width=0.55\linewidth]{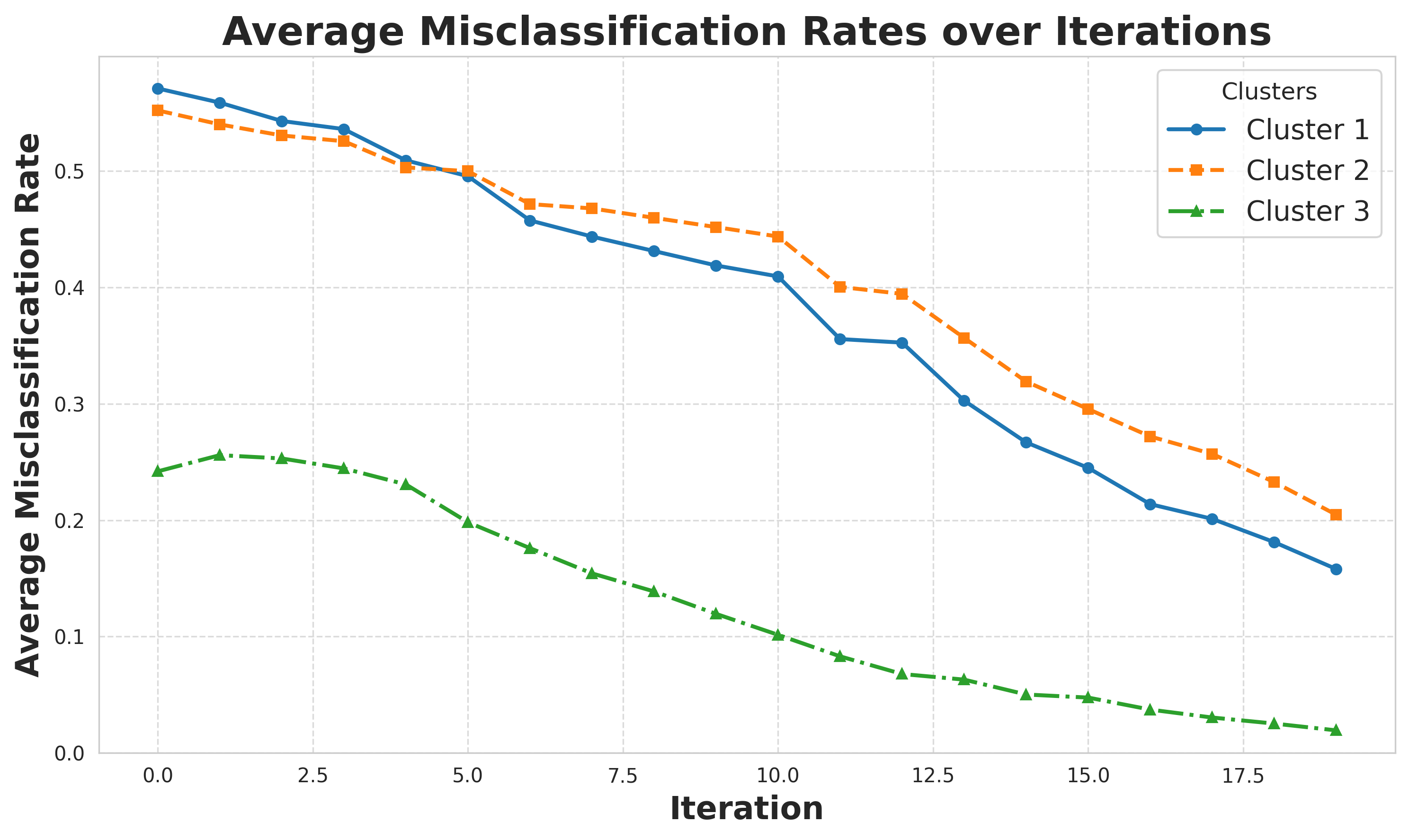}
    \caption{The averages of the misclassification rates over 25 simulations for the $\mathcal{D}_k$'s over iterations.}
    \label{fig:synth1}
\end{figure}
\begin{table}
    \centering
    \caption{Average Misclassification Percentages for PBC and K-Means}
    \label{tab:khat_ablation}
    {
    \begin{tabular}{|c|c|c|c|c|c|c|}
    \hline
        \textbf{$\hat{K}$} & 3 & 4 & 5 & 6 & 7 & 8\\
        \hline
         \textbf{PBC-MLP} & 7.633 & 10.36 & 12.18 & 13.29 & 13.76 &13.65\\
         \textbf{K-Means} & 59.96& 73.11&76.04& 82.63& 85.12& 87.04\\
        \hline
    \end{tabular}}
\end{table}
To assess the performance of our algorithm in the offline setting, i.e., without the ensemble outputs, we compare the cluster assignments of our algorithm to the known ground truth, i.e., which relation generated each point. We calculate the misclassification rate for a simulation as $\frac{1}{N} \sum_{i=1}^{N} \mathbbm{1}_{\{\hat{\gamma}_i \neq \gamma_i\}} $, where $N$ is the total number of data points, $\gamma_i$ is the ground truth cluster of the $i^\text{th}$ data point, $\hat{\gamma}_i$ is the predicted cluster. We assign each predicted cluster a label via majority ground truth, then average misclassification rates over simulations. As shown in Table \ref{tab:khat_ablation}, when $\hat{K}>K$, our misclassification rate does not change significantly, which shows our algorithm is able to cluster points from same mechanisms under different conditions. Further, we examined randomly selected different initializations for $\hat{K}=3$ across 25 simulations; our algorithm yields an average misclassification rate of 7.507\% (standard deviation 3.589\%), demonstrating robustness to initial conditions. We selected $\zeta$ as a reasonable value using our insights from the real-life experiments.

Although our algorithm significantly outperforms the baseline, some data points are still misclassified due to cluster initialization. Initially, clusters are formed from randomly selected points, causing their ML models to learn and remain averages of different mechanisms across iterations. As shown in Fig. \ref{fig:synth1}, the first and second clusters misclassify data from the other cluster by averaging mechanisms, whereas the third cluster, better initialized by selecting points dissimilar to those in the first two clusters, achieves a much lower misclassification rate.

\vspace{-0.5cm}
\subsection{Real Data Experiments}
In this section, we present experiments conducted on the well-studied real-world datasets: the Chicago Crime dataset \cite{crime_kaggle}, the M4 competition dataset \cite{m4}, and other datasets in fields such as finance, climate and healthcare \cite{regression_datasets}, \cite{delhi}.

For these datasets, we compare our Performance Based Clustering (PBC) approach using an MLP model against a vanilla MLP. We construct features from autoregressive and rolling statistical measures and split the data into 80\% training, 10\% validation, and 10\% test sets. The M4 datasets were normalized to [0, 1]. We select hyperparameters for MLP and PBC, such as learning rate, hidden layer sizes, $\zeta$, $\alpha$, and $\hat{K}$, using the validation set. For testing, we use the setting in \ref{ssec:streaming}. We divide the test split into batches $\mathcal{B}^t$ with $N^t$ selected according to the dataset size. We produce outcomes using only $\bm{x}_i^t$'s, then calculate the errors and update the weights using $y_i^t$'s, i.e., data arrives sequentially in a streaming manner.

\begin{figure}
    \centering
    \includegraphics[width=0.55\linewidth]{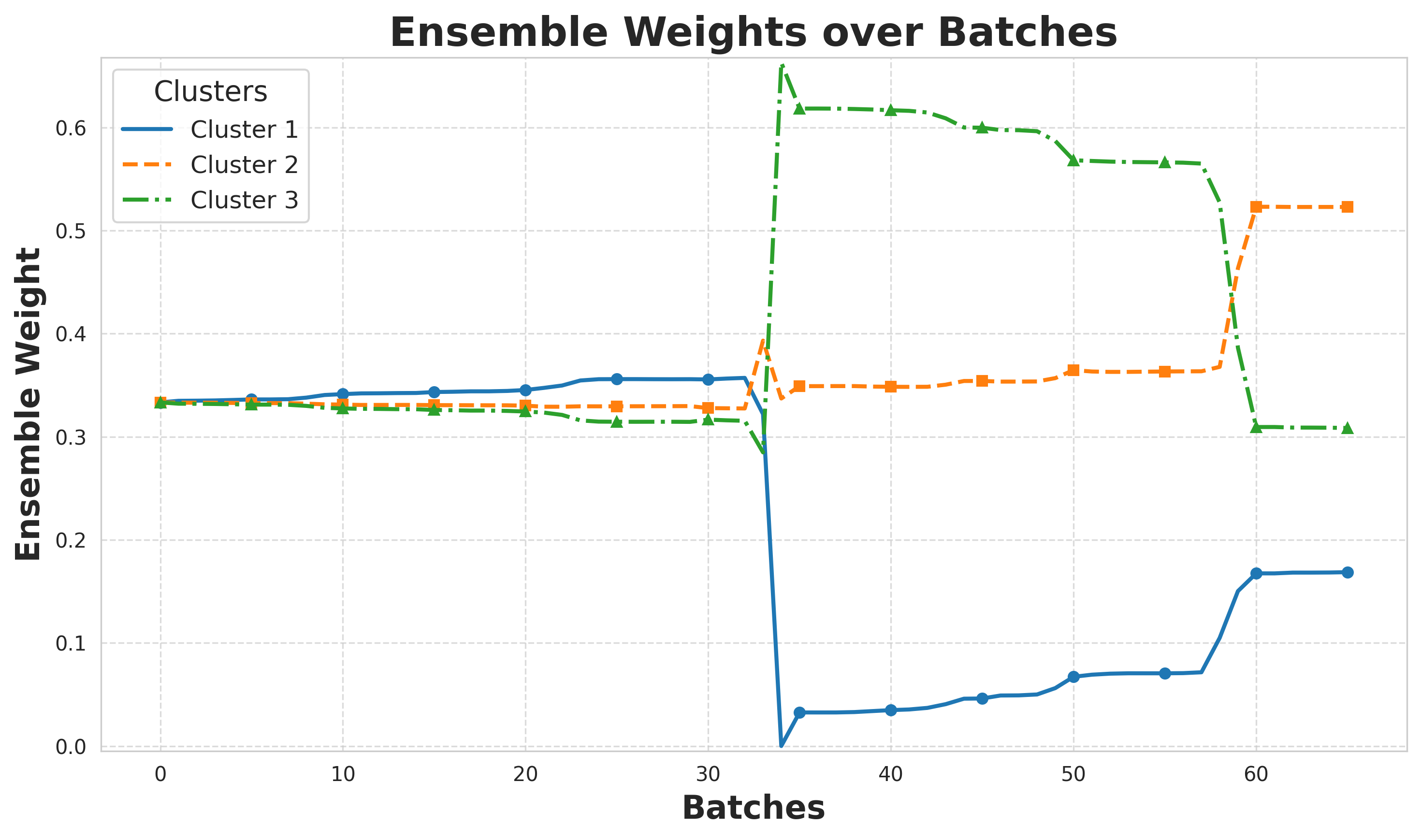}
    \caption{The ensemble weights of the clusters with respect to the sequentially arriving test batches for the M4 Weekly dataset.}
    \label{fig:real1}
\end{figure} 

\vspace{-0.55cm}
\begin{table}[h!]
    \centering
    \caption{Average MSE's for Real-life Datasets}
    \label{tab:results_combined}
    \small 
    \setlength{\tabcolsep}{3pt} 
    \renewcommand{\arraystretch}{0.9} 
    {
    \begin{minipage}{0.48\textwidth} 
        \centering
        \begin{tabular}{{|l|c|c|c|c|c|}}
        \hline
             & \textbf{Crime} & \textbf{M4 Daily} & \textbf{M4 Hourly} & \textbf{M4 Weekly} & \textbf{Stocks  } \\
            \hline
             \textbf{Base MLP} & 3.1127 & 0.001295 & 0.003619 & 0.06908  & 0.005273\\
             \textbf{PBC-MLP} & \textbf{2.9155} & \textbf{0.000847} & \textbf{0.003294} & \textbf{0.02412} & \textbf{0.002367} \\
             \hline
        \end{tabular}
    \end{minipage}%
    \hspace{0.04\textwidth} 
    \begin{minipage}{0.48\textwidth} 
        \centering
        \begin{tabular}{{|l|c|c|c|c|c|}}
        \hline
             & \textbf{Bank8} & \textbf{Bank32} & \textbf{Climate} & \textbf{Diabetes} & \textbf{\;\;\;\;\;\;\;Gas\;\;\;\;\;\;\;} \\
            \hline
             \textbf{Base MLP} & 0.001785 & 0.007928 & \textbf{0.008454} &0.1987 & 0.0009514 \\
             \textbf{PBC-MLP} & \textbf{0.001111} & \textbf{0.006938} & 0.008613 & \textbf{0.03259} & \textbf{0.0007149} \\
             \hline
        \end{tabular}
    \end{minipage}}
\end{table}

In Fig. \ref{fig:real1}, the ensemble weights change significantly at the $34^\text{th}$ and $57^\text{th}$ batches before stabilizing. This indicates that the ensemble of learned functions assigns different weights to various data regions, allowing PBC to adapt to changing patterns in the generating mechanisms, which is beneficial for applications like forecasting non-stationary time series in M4.

Table \ref{tab:results_combined} shows the average MSEs for ten real-world datasets, demonstrating that our approach enhances prediction performance across these datasets. This improvement is due to effectively learning generating mechanisms with separate ML models and the significant contribution of ensemble weight updates, which leverage distinct ML models sequentially with appropriate weights for different data regions.

\vspace{-0.35cm}
\section{Conclusion}\label{sec:conc}
We introduced a clustering approach based on the relations between the feature vectors and the targets, addressing the limitations of the single mechanism assumption. Our algorithm jointly optimizes the cluster assignments and the learned ML models. We extended this to online settings using a weighted ensemble method for applications working with streaming data. Experiments on the well-studied competition datasets demonstrate significant performance improvements over the traditional approaches. Our performance based clustering approach shows superior performance by identifying and modelling distinct data-generating mechanisms in complex and multi-mechanism scenarios.

\clearpage
\bibliographystyle{ieeetr}
\bibliography{refs}

\begin{thebibliography}{10}

\bibitem{wind_kaggle}
B.~Erisen, ``Wind turbine scada dataset.'' Kaggle, 2018.

\bibitem{crime_kaggle}
{City of Chicago}, ``Chicago crime.'' Kaggle, 2018.

\bibitem{clustering1}
M.~Rahmani and G.~K. Atia, ``Subspace clustering via optimal direction search,'' {\em IEEE Signal Processing Letters}, vol.~24, pp.~1793--1797, Dec. 2017.

\bibitem{clustering2}
X.~Ye, S.~Luo, and J.~Zhao, ``Deep bayesian sparse subspace clustering,'' {\em IEEE Signal Processing Letters}, vol.~28, pp.~1888--1892, 2021.

\bibitem{VWKM}
G.~He, W.~Jiang, R.~Peng, M.~Yin, and M.~Han, ``Soft subspace based ensemble clustering for multivariate time series data,'' {\em IEEE Transactions on Neural Networks and Learning Systems}, vol.~34, no.~10, pp.~7761--7774, 2023.

\bibitem{SDCN}
D.~Bo, X.~Wang, C.~Shi, M.~Zhu, E.~Lu, and P.~Cui, ``Structural deep clustering network,'' {\em CoRR}, vol.~abs/2002.01633, 2020.

\bibitem{learningrepresentation}
Z.~Huang, J.~Chen, J.~Zhang, and H.~Shan, ``Learning representation for clustering via prototype scattering and positive sampling,'' {\em IEEE Transactions on Pattern Analysis and Machine Intelligence}, vol.~45, no.~6, pp.~7509--7524, 2023.

\bibitem{SimCLR}
T.~Chen, S.~Kornblith, M.~Norouzi, and G.~E. Hinton, ``A simple framework for contrastive learning of visual representations,'' {\em CoRR}, vol.~abs/2002.05709, 2020.

\bibitem{DeepCluster}
M.~Caron, P.~Bojanowski, A.~Joulin, and M.~Douze, ``Deep clustering for unsupervised learning of visual features,'' {\em CoRR}, vol.~abs/1807.05520, 2018.

\bibitem{DTC}
N.~S. Madiraju, S.~M. Sadat, D.~Fisher, and H.~Karimabadi, ``Deep temporal clustering : Fully unsupervised learning of time-domain features,'' {\em CoRR}, vol.~abs/1802.01059, 2018.

\bibitem{deepcl}
M.~Yue, Y.~Li, H.~Yang, R.~Ahuja, Y.-Y. Chiang, and C.~Shahabi, ``{ DETECT: Deep Trajectory Clustering for Mobility-Behavior Analysis },'' in {\em 2019 IEEE International Conference on Big Data (Big Data)}, (Los Alamitos, CA, USA), pp.~988--997, IEEE Computer Society, Dec. 2019.

\bibitem{em}
T.~K. Moon, ``The expectation-maximization algorithm,'' {\em IEEE Signal Processing Magazine}, vol.~13, pp.~47--60, 1996.

\bibitem{ts1}
X.~L. Li, ``Convolutional pca for multiple time series,'' {\em IEEE Signal Processing Letters}, vol.~27, pp.~1450--1454, 2020.

\bibitem{early_stop}
R.~O. Duda, P.~E. Hart, and D.~G. Stork, {\em Pattern Classification}.
\newblock Hoboken, NJ, USA: Wiley, 2~ed., 2000.

\bibitem{univ_approx}
A.~Heinecke, J.~Ho, and W.-L. Hwang, ``Refinement and universal approximation via sparsely connected relu convolution nets,'' {\em IEEE Signal Processing Letters}, vol.~27, pp.~1175--1179, 2020.

\bibitem{cr_bound}
T.~Huang, Y.~Liu, H.~Meng, and X.~Wang, ``Adaptive compressed sensing via minimizing cramer–rao bound,'' {\em IEEE Signal Processing Letters}, vol.~21, no.~3, pp.~270--274, 2014.

\bibitem{chap-rob}
D.~G. Chapman and H.~Robbins, ``Minimum variance estimation without regularity assumptions,'' {\em The Annals of Mathematical Statistics}, pp.~581--586, 1951.

\bibitem{mct}
S.~Abbott {\em et~al.}, {\em Understanding Analysis}, vol.~2.
\newblock Springer, 2001.

\bibitem{assignment}
C.~Hennig, ``Identifiablity of models for clusterwise linear regression,'' {\em Journal of classification}, vol.~17, 2000.

\bibitem{dl}
I.~J. Goodfellow, Y.~Bengio, and A.~Courville, {\em Deep Learning}.
\newblock Cambridge, MA, USA: MIT Press, 2016.
\newblock \url{http://www.deeplearningbook.org}.

\bibitem{m4}
S.~Makridakis, S.~Evangelos, and A.~Vassilios, ``The m4 competition: Results, findings, conclusion and way forward,'' {\em International Journal of forecasting}, vol.~34, no.~4, pp.~802--808, 2018.

\bibitem{regression_datasets}
L.~Torgo, ``Regression datasets.'' \url{http://www.dcc.fc.up.pt/~ltorgo/Regression/DataSets.html}.
\newblock Accessed: Jun. 15, 2023.

\bibitem{delhi}
S.~Rao, ``Daily climate time series data.'' \url{https://www.kaggle.com/datasets/sumanthvrao/daily-climate-time-series-data/data}, 2020.
\newblock Kaggle.

\end{thebibliography}

\ifCLASSOPTIONcaptionsoff
  \newpage
\fi

\end{document}